\pdfoutput=1
\documentclass[11pt]{article}

\usepackage[]{ACL2023}
\usepackage{times}
\usepackage{latexsym}
\usepackage{epsfig}
\usepackage{graphicx}
\usepackage{amsmath}
\usepackage{amssymb}
\usepackage{algpseudocode}

\usepackage{multirow}
\usepackage{booktabs}
\usepackage{algorithm}
\usepackage{subfigure}
\usepackage{bbm}
\usepackage{amssymb}
\usepackage{colortbl}
\definecolor{mypink}{rgb}{.99,.91,.95}
\definecolor{mygreen}{rgb}{.9,.99,.9}
\definecolor{mygray}{gray}{.9}

\usepackage{tcolorbox}
\usepackage{listings}

\usepackage[T1]{fontenc}
\usepackage[utf8]{inputenc}

\usepackage{microtype}

\usepackage{inconsolata}

\title{E$^{2}$-LLM: Efficient and Extreme Length Extension of  \\Large Language Models}
\author{Jiaheng Liu*$^{1}$, Zhiqi Bai*$^{1}$, Yuanxing Zhang$^{1}$, Chenchen Zhang$^{1}$, \textbf{Yu Zhang}$^{1}$, \\  \textbf{Ge Zhang}$^{2}$, \textbf{Jiakai Wang}$^{1}$,  \textbf{Haoran Que}$^{1}$, \textbf{Yukang Chen}$^{3}$, \textbf{Wenbo Su}$^{1}$, \textbf{Tiezheng Ge}$^{1}$, \\ \textbf{Jie Fu}$^{4}$, \textbf{Wenhu Chen}$^{2}$, \textbf{Bo Zheng}$^{1}$\\
       $^{1}$Alibaba Group, $^{2}$University of Waterloo, $^{3}$The Chinese University of Hong Kong, \\
       $^{4}$The Hong Kong University of Science and Technology \\
        \texttt{\{ljh411989, baizhiqi.bzq\}@taobao.com}}

\begin{document}

\maketitle
\let\thefootnote\relax\footnotetext{* First two authors contributed equally.}
\begin{abstract}
% We present an Arbitrary Position Interpolation (API) strategy, an efficient fine-tuning approach that extends the context sizes of pre-trained large language models (LLMs), with limited computation cost. 
Typically, training LLMs with long context sizes is computationally expensive, requiring extensive training hours and GPU resources.
Existing long-context extension methods usually need additional training procedures to support corresponding long-context windows,
where the long-context training data (\textit{e.g.}, 32k) is needed, and high GPU training costs are assumed.
To address the aforementioned issues,
we propose an Efficient and Extreme length extension method for Large Language Models,
% Interpolation for eXtreme Length extension method, 
called E$^{2}$-LLM, with only one training procedure and dramatically reduced computation cost,
which also removes the need to collect long-context data.
Concretely,
first,
the training data of our E$^{2}$-LLM only requires a short length (\textit{e.g.}, 4k),
which reduces the tuning cost greatly.
Second,
the training procedure on the short training context window is performed only one time, and we can support different evaluation context windows at inference.
Third,
in E$^{2}$-LLM,
based on RoPE position embeddings,
we introduce two different augmentation methods on the scale and position index parameters for different samples in training. 
% we introduce to randomly select the scale parameter proposed in PI (Position Interpolation) method and the starting position index in RoPE for each sample,
It aims to make the model more robust to the different relative differences when directly interpolating the arbitrary context length at inference.
Comprehensive experimental results on multiple benchmark datasets demonstrate the effectiveness of our E$^{2}$-LLM on challenging long-context tasks.
% Furthermore,
% our method also preserves the short-context understanding ability well when compared to existing extension methods.
% where short-context understanding ability usually degrades for existing extension methods.
% For example, training on the context length of 8192 needs 16× computational costs in self-attention layers as that of 2048. To achieve this goal, Position Interpolation linearly down-scales the input position indices to match the original context window size, rather than extrapolating beyond the trained context length which may lead to catastrophically high attention scores that completely ruin the self-attention mechanism. Our theoretical study shows that the upper bound of interpolation is at least smaller than that of extrapolation, further demonstrating its stability. Models extended via Position Interpolation retain their original architecture and can reuse most pre-existing optimization and infrastructure.
\end{abstract}

\section{Introduction}
\label{sec:intro}
% \clj{In recent years, large language models (LLMs) have surpassed expectations, with notable examples like Chat-GPT, GPT-4 from OpenAI, and LLaMA from Meta. These models have demonstrated impressive capabilities in neutral language processing tasks. A key factor contributing to their success is the incorporation of long context windows, enabling the models to grasp contextual information effectively.
% The inclusion of long context windows empowers LLMs to excel in long-form generation tasks. This capability unlocks a wide range of practical applications, such as incorporating extended conversational context, writing or continuing novels, summarizing financial or legal reports, and providing detailed answers to important questions.}   \yuandong{We should start from why long-form language model tasks are important and list a few existing models}. 
% }

Large language models (LLMs) usually have a pre-defined context window length. 
For example, inputs to LLaMA models~\citep{llama,llama2} must be fewer than 2,048 or 4096 tokens. 
This pre-set context window limit is frequently exceeded in applications such as long conversations, document summarization, or long-term reasoning~\cite{zheng2023judging,position-interpolation}.
For these applications, LLMs with longer context windows are preferred. 
However, training an LLM from scratch with long context windows requires significant training costs. 
% Recently,
To address this problem,
many long-context extension methods~\cite{yarn,longlora} have been proposed to
% This raises a question: Can we
extend the context window of an existing pre-trained LLM.

\begin{figure}[t]
\begin{center}
\includegraphics[width=1.0\linewidth]{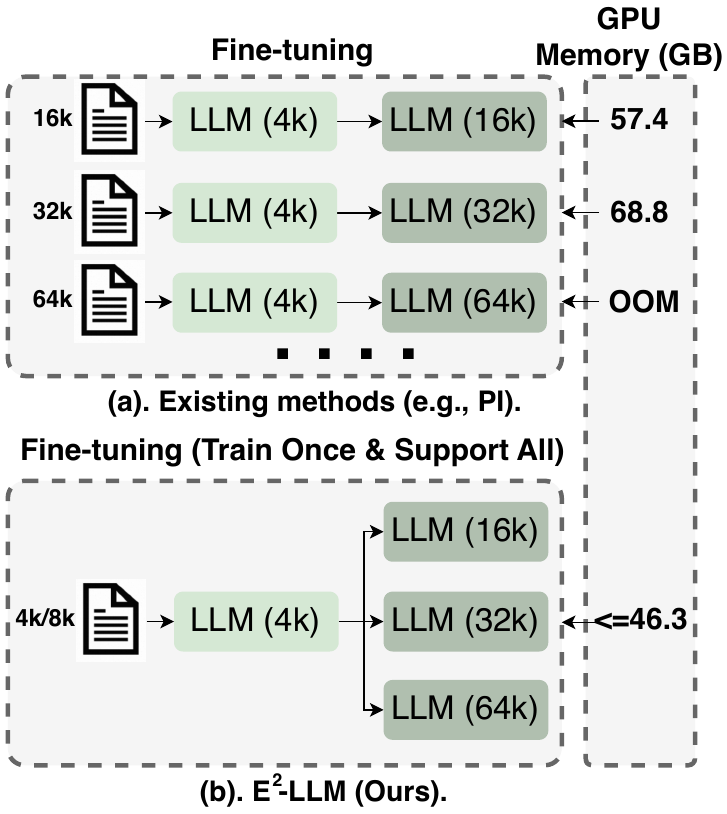}
% \vspace{-5mm}
\caption{Comparison of existing long-context extension methods (\textit{e.g.}, PI~\cite{position-interpolation}) and our E$^{2}$-LLM. ``LLM (4k)'' with light color means the LLM with default context window (\textit{e.g.}, LLaMa2 with 4k). ``LLM (16k/32k/64k)'' with deep color means the LLM with extended context windows (16k/32k/64k) after fine-tuning. (a) For existing methods, we need to collect long-context data (\textit{e.g.}, 16k/32k) and fine-tune the LLM models for different context extension windows with high GPU memory usage.
(b). For E$^{2}$-LLM, we directly use the short-context data (\textit{e.g.}, 4k/8k) by training LLMs only once time with acceptable GPU memory usage and support different evaluation context windows (\textit{e.g.}, 16k/32k/64k) at inference.}
\label{fig:intro}
% \vspace{-6mm}
\end{center}
\end{figure}
One straightforward approach is called direct extrapolation,
which fine-tunes an existing pre-trained LLM with a longer context window,
% trained this way adapt to long context windows very slowly.
% which is 
However, the authors of Position Interpolation (PI)~\cite{position-interpolation} observe that models trained by direct extrapolation adapt to long context windows very slowly and direct extrapolation is inefficient in extending substantially longer context windows.
% Many long-context extension methods have recently been proposed~\cite{position-interpolation,yarn,longlora}.
% \ignore{
% \begin{figure}[]
% \centering
% \includegraphics[width=0.8\textwidth]{figure-1.pdf}
% \caption{
% \small Effective context window size of LLaMA after extending using direct fine-tuning v.s. Position Interpolation. 
% Direct fine-tuning method increases the effective context
% window size very slowly, only to slightly more than the original 2048 after fine-tuining for 1000 steps.
% Position Interpolation can enable full effective context window after tuning within 100 steps. 
% See Section~\ref{sec:passkey} for the measurement method of effective context window size and Section~\ref{sec:setup} for 
% training setup.
% }
% \label{fig:effective-window-size}
% \end{figure}
% }

% In this work, 
% we present the Arbitrary Position Interpolation (\textbf{API}) strategy to enable arbitrary context window extensions for certain existing pre-trained LLMs, including LLaMA with limited computation cost. Typically, training LLMs with long context sizes are computationally expensive, requiring extensive training hours and GPU resources.
As shown in Fig.~\ref{fig:intro} (a),
existing long-context extension methods (\textit{e.g.}, PI) usually need additional training procedures to support corresponding longer-context windows,
where the long-context training data is needed to collect, and training with high GPU memory usage is needed for each context window\footnote{Note that we follow~\cite{longlora} to report the GPU memory by fine-tuning LLaMA2 7B on various context lengths with FlashAttention-2~\cite{flash-attention2} and DeepSpeed stage 2~\cite{deepspeed}.}.

To address the aforementioned issues,
as shown in Fig.~\ref{fig:intro} (b),
we propose an Efficient and Extreme length extension method of LLMs, called \textbf{E$^{2}$-LLM}, to support the extreme length extension of LLMs with only one training procedure on short-context data and limited computation cost.
Based on E$^{2}$-LLM,
we can support different evaluation context windows well at inference by only changing one hyperparameter of RoPE~\cite{rope} according to the input context length.
% and the collection of long-context data (\textit{e.g.}, 16k/32k) is also not needed.
Specifically,
first,
in E$^{2}$-LLM,
the training data only requires the commonly-used data with short lengths (\textit{e.g.}, 4k), and we only need to train the LLM once and support interpolating the arbitrary context length at inference,
which reduces the GPU usage costs largely.
Second,
in our E$^{2}$-LLM,
% our E$^{2}$-LLM includes two variants (i.e., the initial version called \textbf{API-Ini} and the better version called \textbf{API-Pro}).
% % the training data of our API only requires the commonly-used data with short lengths (\textit{e.g.}, 2k/4k),
% % which reduces the efforts of collecting long-context training data.
% % Second,
% For  API-Ini,
% in our API,
we first propose the augmentation on the scale parameter of PI from a predefined distribution (\textit{e.g.}, uniform distribution),
which aims to cover different position densities in training.
% which aims to make our LLM more robust 
Besides,
we observe that only changing the scale parameter will make LLMs focus on a small range of absolute position indices. 
Therefore,
in our E$^{2}$-LLM,
% and the starting position index in RoPE for each sample,
% For API-Pro,
to improve the generalization ability of our E$^{2}$-LLM,
we further propose the augmentation on the position index parameters by introducing the position offsets on the absolute position indices of RoPE.
% To make the model more robust to the relative difference and ignore the influences of absolute position values.
% we propose to randomly select the starting position index for RoPE to train the different absolution position values.
% which aims 

% Third,
% for our API strategy,

The contributions of our E$^{2}$-LLM are  as follows:
\begin{itemize}
    \item In our work, we first investigate the issues (\textit{e.g.}, the large fine-tuning costs with long context data) of existing long-context extension methods, and propose the Efficient and Extreme length extension method (i.e., \textbf{E$^{2}$-LLM}) to train the LLMs once on the short-context data with limited GPU memory costs and support different evaluation context windows.
    % analyze the influence of the existing face distillation method from the distribution of positive pairs and negative pairs. The former represents intra-class compactness, and the latter represents the distribution between classes.
    \item  In E$^{2}$-LLM, based on RoPE, we propose two augmentation strategies on the scale and position index parameters for different samples in training,
    which aims to support different evaluation context windows within one training procedure and improve the long-context abilities of LLMs.
    % \item In our E$^{2}$-LLM, we first provide the initial version (\textbf{API-Ini}) by randomly selecting the scale parameter proposed in position interpolation and then introduce the better version (\textbf{API-Pro}) by randomly changing the starting position index of RoPE to make the LLMs more robust to the relative differences and ignore the absolute position index values.
    \item  
Comprehensive experimental results on multiple long-context benchmark datasets demonstrate the effectiveness and efficiency of our E$^{2}$-LLM method.
% \item 
% % Besides,
% Our method also preserves the short-context understanding ability well when compared to existing long-context extension methods,
% where short-context understanding ability usually degrades for existing methods.
\end{itemize}

\section{Related Works}

\paragraph{Long-context Transformers.}
Extensive studies have aimed to increase transformers' ability to process longer text sequences. Strategies like using retrieval-based models \citep{retrieval-QA, few-shot-retrieval} have been employed, which integrate additional documents and search findings into the context. 
% Our research enhances these methods by preserving the original attention process during inference.
Various efforts have adapted the multi-head attention by devising estimated alternatives \citep{linformer,long-former, reformer, recurrent-memory-transformer, longnet} to mitigate the self-attention's inherently high computational demands. For example, Longformer \citep{long-former} and BigBird \citep{big-bird} use a form of diluted attention for more extensive text. Meanwhile, other initiatives \citep{memorizing-transformer, recurrent-memory-transformer} have introduced memory-based systems to condense previous inputs and recall pertinent components. However, these approaches tend to fall short of the effectiveness of complete attention, thereby hindering the refinement of large pre-trained language models (LLMs)~\cite{concepthmath,guo2023owl,wang2023rolellm,mtbench,chai2024xcot}.
Our approach differs in that it approximates the attention mechanism in a way that remains closely aligned with the conventional attention method, showing only a minimal discrepancy.

\paragraph{Long-context LLMs.}
Large language models (LLMs) such as LLaMA \citep{llama} and LLaMA2 \citep{llama2} are originally trained with fixed context sizes, typically 2048 and 4096 tokens, respectively.
Nonetheless, the cost of training LLMs with extended contexts from the ground up is usually beyond the reach of the average research team. Consequently, recent studies have explored ways to expand the context length of these models during the fine-tuning stage. For example, Position Interpolation \citep{position-interpolation} adapts the rotary position encoding technique \citep{rope}, which allows LLaMA to process contexts as long as 32768 tokens.
% Similarly, the Focused Transformer \citep{focused-transformer} leverages contrastive learning in traning.
Another method, Landmark attention \citep{landmark-attention}, achieves efficiency but at the cost of some accuracy, by compressing extended contexts into a set of retrieved tokens. In contrast, our strategy minimizes the expenses related to fine-tuning without compromising the efficacy of the original attention. It ensures that the model has complete and unchanged attention over the full input during the inference process.
Other approaches, like ALiBi \citep{alibi}, have been designed to train Transformers on shorter sequences and then apply them to longer ones at inference time, effectively extrapolating context length. However, these techniques are not as effective for pre-trained LLMs that rely on positional encodings with poor extrapolation capabilities, such as RoPE \citep{rope}. To overcome this, recent research has been directed towards modifying the positional embeddings of LLMs to handle longer texts. This includes methods like Position Interpolation \citep{position-interpolation}, NTK-aware position embeddings \citep{ntk-pe}, and Yarn \citep{yarn}.

\section{Background}
\label{back}
\def\vf{\mathbf{f}}
\def\vx{\mathbf{x}}
\def\vk{\mathbf{k}}
\def\vq{\mathbf{q}}
\def\vu{\mathbf{u}}
\def\di{\mathrm{i}}
\subsection{Rotary Position Embedding (RoPE)} 
Transformer models require explicit positional information to be injected,
where the positional encodings are used to represent the order of inputs. 
In this section,
we take Rotary Position Embedding (RoPE) \citep{rope} as an example, which is widely-used in many LLaMA-style models \citep{llama}.
In RoPE, given a position index $m \in [0, L)$ and an embedding vector $\vx := [x_0, x_1, \ldots, x_{d-1}]^\top$, where $L$ is the context window and $d$ is the dimension of the attention head, a vector-valued complex function $\vf(\vx, m)$ is defined as follows:
\begin{equation}
\small
    \vf(\vx,m) = [(x_0 + \di x_1) e^{\di m \theta_0}, \ldots, (x_{d-2} + \di x_{d-1})e^{\di m \theta_{d/2-1}}]^\top,
    \label{eq:rope2}
\end{equation}
where $\di := \sqrt{-1}$ is the imaginary unit and $\theta_j = 10000^{-2j/d}$. Based on RoPE, the self-attention score $a$ is computed as follows:
% \begin{eqnarray}
% a(m,n) &=& \mathrm{Re}\langle\vf(\vq, m), \vf(\vk, n)\rangle \nonumber \\
% &=& \mathrm{Re}\left[\sum_{j=0}^{d/2-1} (q_{2j} +\di q_{2j+1})(k_{2j} - \di k_{2j+1}) e^{\di (m-n)\theta_j}\right] \nonumber \\
% &=& \sum_{j=0}^{d/2-1} (q_{2j} k_{2j} + q_{2j+1}k_{2j+1})\cos((m-n)\theta_j) + (q_{2j} k_{2j+1} - q_{2j+1}k_{2j})\sin((m-n)\theta_j) \nonumber \\
% &=:& a(m-n) \label{eq:attn-score}
% \end{eqnarray}
\begin{eqnarray}
% \small
a(m,n) &=& \mathrm{Re}\langle\vf(\vq, m), \vf(\vk, n)\rangle \nonumber \\
% &=& \mathrm{Re}\left[\sum_{j=0}^{d/2-1} (q_{2j} +\di q_{2j+1})(k_{2j} - \di k_{2j+1}) e^{\di (m-n)\theta_j}\right] \nonumber \\
 &=:&a(m-n) \label{eq:attn-score},
\end{eqnarray}
where $\vq$ and $\vk$ are the query and key vectors for a specific attention head, respectively, and the detailed derivation is omitted.
In Eq.~\ref{eq:attn-score},
we observe that $a(m,n)$
is only dependent on relative position $m-n$ through trigonometric functions.  Besides, RoPE is performed on both query and key embeddings for calculating the attention scores at each layer. 

\iffalse
\begin{equation}
	\vf(\vx, m) = 
	\begin{pmatrix}
		x_1\\
		x_2\\
		x_3\\
		x_4\\
		\vdots\\
		x_{d-1}\\
		x_d
	\end{pmatrix}
	\cdot
	\begin{pmatrix}
		\cos{m\theta_1} \\
		\cos{m\theta_1} \\
		\cos{m\theta_2} \\
		\cos{m\theta_2} \\
		\vdots \\
		\cos{m\theta_{d/2}} \\
		\cos{m\theta_{d/2}} 
	\end{pmatrix}
	+
	\begin{pmatrix}
		-x_2\\
		x_1\\
		-x_4\\
		x_3\\
		\vdots\\
		-x_d\\
		x_{d-1}
	\end{pmatrix}
	\cdot
	\begin{pmatrix}
		\sin{m\theta_1}\\
		\sin{m\theta_1}\\
		\sin{m\theta_2}\\
		\sin{m\theta_2}\\
		\vdots\\
		\sin{m\theta_{d/2}}\\
		\sin{m\theta_{d/2}}
	\end{pmatrix},
\end{equation}
where $\cdot$ is element-wise multiplication, and $\theta_k = 10000^{-2k/d}$ is a constant. 
\fi

\subsection{Position Interpolation}
While the attention score in Eq.~\ref{eq:attn-score} only depends on the relative positions,
its extrapolation performance is not great. Specifically, when direct extrapolation to larger unseen context windows in training, the perplexity will increase to very high numbers (i.e., $>10^3$).
Recently, Position Interpolation (PI)~\cite{position-interpolation} has been proposed, where $s$ is defined as the positional span between a query and a key, and $L$ is defined as the size of the trained context window.
% which is also the supported extension window.
Instead of direct extrapolation on the attention score to $s > L$, 
the attention score is defined as $\tilde a(s) = a(Ls/L')$,
where $L'$ is the extended longer context window.
Formally,
in PI,
RoPE $\vf$ is replaced by $\vf’$ as follows:
\begin{equation}
    \vf’(\vx, m)= \vf\left(\vx, \frac{mL}{L'} \right),
    \label{pi}
\end{equation}
where position indices from $[0, L')$ to $[0, L)$ are reduced to match the original range of indices before computing RoPE.
In other words, the maximum relative distance between any two tokens has been reduced from $L'$ to $L$ and PI reduces the effect on attention score computation when extending the context window, and makes the LLMs easier to adapt. 
% As a result of being fed into RoPE, the greatest distance that separates any pair of tokens has decreased from $L'$ to $L$. Additionally, Positional Infusion (PI) diminishes the impact that extending the context window has on calculating attention scores, potentially simplifying the model's adaptation process.
Furthermore,
we define the scale parameter $g$ as $\frac{L'}{L}$.
For example,
$g$ is set as $2$ when $L'=8192$ for LLaMa2 with context window of $L=4096$.
Thus,
the Eq.~\ref{pi} can be reformulated as follows:
% To further demonstrate this is the case, in the following theorem, we show that the interpolated attention score is well-behaved: 
\begin{equation}
    \vf’(\vx, m)= \vf\left(\vx, \frac{m}{g} \right).
    \label{pi2}
\end{equation}
Meanwhile,
for PI,
we need to collect the long-context data with the maximum length of $L'$ in finetuning, and finetuning is needed for each extended window with high GPU memory usage as shown in Fig.~\ref{fig:intro} (a).
\section{Method}
% In this section,
In this section, we introduce the details of our E$^{2}$-LLM in Fig.~\ref{fig:intro} (b) for extending different sizes of context windows by only performing one training procedure on short-length data,
which reduces the tuning costs greatly.
% where we apply the augmentation strategies on the 
% which includes two variants (i.e., \textbf{API-Ini}, \textbf{API-Pro})
% in Fig.~\ref{fig:Overview}.
First, in Sec.~\ref{notation},
we provide the necessary notations.
Then, in Sec.~\ref{api},
we illustrate the details of our E$^{2}$-LLM strategy to improve the length extension performance by introducing the augmentation on the scale, and the position offset parameters of RoPE,
where these parameters are defined in Eq.~\ref{pi3}.
% we illustrate our initial API strategy to improve the interpolation performance by randomly choosing the scale parameter defined in PI.
% After that,
% in Sec.~\ref{api-pro},
% based on API,
% we provide the better version (i.e., {API-Pro}) to achieve better performance by randomly choosing the starting position index of the RoPE.
Finally,
in Sec.~\ref{train},
we show the training and inference processes in our E$^{2}$-LLM.
% First, at the training stage, for each sample in the current batch,
% we randomly select the scale parameter $g$ and the starting position index $n$ for the position interpolation.
\subsection{Notations}
\label{notation}
Apart from the notations defined in Sec.~\ref{back},
we also define the following notations.
First, the trained length is defined as $R$.
It should mentioned that $R$ is the maximum length of the data in finetuning,
which is set as 8k in E$^{2}$-LLM, by default.
Therefore,
it is easy to collect the training data with a length of $R$ and the used GPU memory in finetuning is also acceptable. 
In contrast,
the trained length $R$ is equal to the extension length $L'$ (\textit{e.g.}, 16k/32k) in many long-context extension methods (\textit{e.g.}, PI),
which requires high GPU memory usage in training.
Second, we also introduce the  position offset $t$ in RoPE, and  we can reformulate Eq.~\ref{pi2} to compute the RoPE embeddings as follows:
\begin{equation}
    \vf’(\vx, m)= \vf\left(\vx, \frac{m+t}{g} \right).
    \label{pi3}
\end{equation}
In standard RoPE,
by default,
the $t$ is set as $0$.
In our E$^{2}$-LLM, 
the $t$ is selected from a range of indices  $T=\{0, ..., t_{max}\}$,
where $t_{max}$ is the maximum position offset.
Third,
we also define a set of scale parameters used in E$^{2}$-LLM as $G=\{1, 2, ..., g_{max}\}$,
where $g_{max}$ is the maximum scale parameter.
\subsection{E$^{2}$-LLM}
\label{pixl}
In this section, we describe our proposed E$^{2}$-LLM strategy in detail.
We take the LLM model $\mathcal{H}$ with the default context window $L$ as 4,096 and the trained length $R$ as 4,096 for illustration.
We propose two different augmentation methods on the hyperparameters (\textit{i.e.}, the scale parameter $g$ and the position offset $t$) of RoPE.
\subsubsection{Augmentation on $g$}
\label{api}

\begin{figure}[t]
\begin{center}
\includegraphics[width=1.0\linewidth]{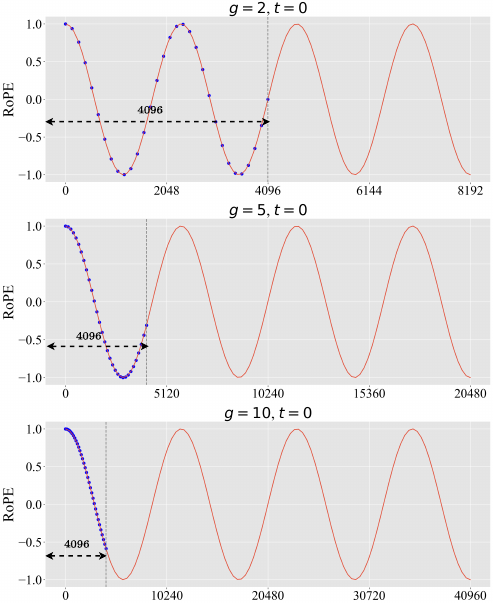}
% \vspace{-5mm}
\caption{The trained position indices (blue points) when using different scale parameters (i.e., $g=2,5,10$). The maximum length of the finetuning data (i.e., $R$) is 4096 and the position offset $t$ is set as $0$ for illustration.}
\label{fig:api}
% \vspace{-6mm}
\end{center}
\end{figure}

% \subsubsection{Augmentation on $g$.}
As shown in Fig.~\ref{fig:api},
we illustrate the augmentation procedure on the scale parameter $g$.
% in training,
% for each sample, to make the 

In the training process of our proposed E$^{2}$-LLM,
 to make the model $\mathcal{H}$ cover different position densities in training,
for the $i$-th iteration, 
we sample the scale parameter $g_i$ from $G$ for different iterations following a predefined probability distribution $P$ as follows:
\begin{equation}
   g_i = \mathcal{S}_g(P, G), g_i\in G,
    \label{eq:scale}
\end{equation}
where $\mathcal{S}_g(P, G)$ denotes the sampling operation on $g$, which samples $g_i$ from set $G$ following the distribution $P$.
Therefore, different scale parameters are used for different iterations.
Note that  $P$ is based on uniform distribution, by default.

In Fig.~\ref{fig:api},
we set the position offset $t$ as 0,
and then randomly select the scale parameter $g$ from $G$ for each sample based on Eq.~\ref{pi3},
where  $g$ is set as 2, 5, and 10, respectively.
Besides,
as shown in Fig.~\ref{fig:api},
we observe that the interpolated maximum context windows are different for different samples in training,
and the densities of the trained position indices are different.
% Besides,
% we set the starting position index $t$ as 0,
% and then randomly select the scale parameter $g$ from $G$ for each sample based on Eq.~\ref{pi2},
% where the $g=2, 5, 10$ in Fig.~\ref{fig:api}.
For example, 
the interpolated interpolated context windows are 8,192 and 20,480 when $g$ is 2 and 5,
respectively.
Furthermore,
as the training context window $R$ is less than the interpolated maximum context windows,
only a certain proportion of the position indices are trained,
which are shown in blue in Fig.~\ref{fig:api}.

% In Fig.~\ref{fig:api},
% we visualize the training position indices shown in blue points by using different numbers of scale parameters (i.e., $g=2, 5, 10$).
% Specifically,
% in Fig.~\ref{fig:api},
% the blue points are the trained position indices by setting different scale parameters,
% and the red position indices are the originally trained positions in  $\mathcal{H}$.
% Furthermore,
% by setting different $g$,
% and we observe that our API-Ini can  cover different position densities in training,
% \jh{which aims to make our model $\mathcal{H}$ focus on the relative distances of different positions. (Why)}

\begin{figure}[t]
\begin{center}
\includegraphics[width=1.0\linewidth]{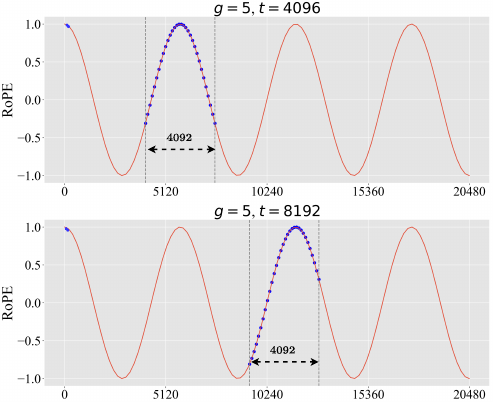}
% \vspace{-5mm}
\caption{The trained position indices (blue points) when using different position offsets and $g$ is set as $5$ for visualization. The position indices of the first four tokens are not moved.}
\label{fig:api-pro}
% \vspace{-6mm}
\end{center}
\end{figure}
% \begin{equation}
%     \vf’(\vx, m)= \vf\left(\vx, \frac{mL}{L'} \right),
%     \label{pi}
% \end{equation}
% \begin{figure}[t]
% % \begin{center}
%     \centering
%     \subfigure{
%         \includegraphics[width=1.0\linewidth]{API_n2.pdf}
%         \label{label_for_cross_ref_1}
%     }\vspace{-3mm}
%     \subfigure{
% 	\includegraphics[width=1.0\linewidth]{API_n5.pdf}
%         \label{label_for_cross_ref_2}
%     }\vspace{-3mm}
%     \subfigure{
%     	\includegraphics[width=1.0\linewidth]{API_n10.pdf}
%         \label{label_for_cross_ref_3}
%     }
% % \includegraphics[width=1\linewidth]{API_n2.pdf}
% % \vspace{-3mm}
% \caption{API-Ini.}
% % source description in English, together with its different .}
% \label{fig:api}
% % \vspace{-5mm}
% % \end{center}
% \end{figure}

\subsubsection{Augmentation on $t$}
\label{api-pro}
As shown in Fig.~\ref{fig:api},
we observe that we can only focus on a small range of the position indices when we start from zero index (i.e., $t=0$).
Therefore,
to improve the robustness and generalization ability of our E$^{2}$-LLM,
we further introduce the augmentation procedure on the position offset $t$ by changing the absolute position indices of RoPE.
% by randomly selecting the starting position index $t$ from $T$.
% Specifically,
% as shown in Fig.~\ref{fig:api-pro},
% and we reformulate Eq.\ref{pi2} to compute the RoPE as follows:
% \begin{equation}
%     \vf’(\vx, m)= \vf\left(\vx, \frac{m+t}{g} \right), t\in T.
%     \label{pi3}
% \end{equation}
% Similarly,
Besides,
inspired by several recent works~\cite{lm-infinite,xiao2023streamingllm},
which claim that a large amount of attention scores are allocated to the initial tokens (i.e., attention sinks~\cite{xiao2023streamingllm}), 
we also keep several initial tokens and set the position offsets of these tokens as 0.
For other position indices,
in the $i$-th training iteration,
we set the  position offset $t$ for different position indices of the trained window as follows:
 
\begin{equation}
   t_i = \begin{cases}
      0,    &  m \in [0, 3]  \\
      \mathcal{S}_t(Q, T), &  m \in (3, R)
   \end{cases},
   \label{eq:augt}
\end{equation}
where $\mathcal{S}_t(Q, T)$ denotes the sampling operation on $t$,
and samples $t_i$ from set $T$ following the predefined probability distribution $Q$.
{Note that $Q$ is set as a uniform distribution and $t_{max}$ is set as the difference between the maximum interpolated context window and the trained context window in the current iteration.}
Based on Eq.~\ref{eq:augt},
for $n \in [0, 3]$ and $m \in (3, R)$,
the Eq.~\ref{eq:attn-score} can be written as follows:
\begin{eqnarray}
\small
a(m,n) &=& \mathrm{Re}\langle\vf(\vq, m+t_i), \vf(\vk, n+t_i)\rangle \nonumber \\
% &=& \mathrm{Re}\left[\sum_{j=0}^{d/2-1} (q_{2j} +\di q_{2j+1})(k_{2j} - \di k_{2j+1}) e^{\di (m-n)\theta_j}\right] \nonumber \\
&=:&a(m+\mathcal{S}_t(Q, T)-n) \label{eq:new-attn-score}.
\end{eqnarray}
Therefore,
when $\mathcal{S}_t(Q, T)$ is larger,
the range of relative position differences (i.e., $(m+\mathcal{S}_t(Q, T)-n)$) between $m$ and $n$ is larger,
which will make the model generalize to different ranges of relative position differences.

% Specifically,
In Fig.~\ref{fig:api-pro},
we also provide the trained position indices (i.e., blue points) when introducing the augmentation on the position offset $t$,
and observe that E$^{2}$-LLM can easily make use of the position indices with different absolute values and different ranges of relative differences.

% \jh{List the maximum length.}
% Meanwhile, inspired by the StreamingLLM~\cite{xiao2023streamingllm}, which claims that a large amount of attention score is allocated to the initial tokens (i.e., attention sinks~\cite{xiao2023streamingllm}), 
% we also keep several initial tokens and  
% \begin{figure}[t]
% % \begin{center}
%     \centering
%     \subfigure{
%         \includegraphics[width=1.0\linewidth]{API_n5_r4096.pdf}
%         \label{label_for_cross_ref_1}
%     }\vspace{-3mm}
%     \subfigure{
% 	\includegraphics[width=1.0\linewidth]{API_n5_r8192.pdf}
%         \label{label_for_cross_ref_2}
%     }
% % \includegraphics[width=1\linewidth]{API_n2.pdf}
% % \vspace{-3mm}
% \caption{API-Pro.}
% % source description in English, together with its different .}
% \label{fig:api-pro}
% % \vspace{-5mm}
% % \end{center}
% \end{figure}

% \subsection{Analysis}
% \jh{Add theories or visualizations to further discuss the advantages of our E$^{2}$-LLM.}
\begin{algorithm}[t]
% \footnotesize
\caption{Training of E$^{2}$-LLM}
\label{alg:pseudocode}
\begin{algorithmic}[1]
\Require
   Pre-trained LLM model $\mathcal{H}$ with default context window of $L$ (\textit{e.g.}, 4k); The trained context window is $R$ (\textit{e.g.}, 4k/8k); The evaluation context window $L'$ (\textit{e.g.}, 32k/64k);
   % Randomly initialized student model $\mathcal{S}$;
  % Classifier $\mathcal{C}$; 
  % Current batch with $N$ samples;
  % Supported scale parameters $G=\{1, 2, ..., g_{max}\}$ ($g_{max}$ is the maximum scale parameter);
  % Range of starting position index  $T=\{1, 2, ..., t_{max}\}$ ($t_{max}$ is the maximum starting position index).
  % The feature dimension $d$;
  % The maximum number of features for each identity $K$;
  % The number of identities $Q$;
  % The student feature bank $\mathbf{M}^{s}\in \mathbb{R}^{Q\times K\times d}$ and  teacher feature bank $\mathbf{M}^{t}\in \mathbb{R}^{Q\times K\times d}$;
  % $\mathbf{S}\in \mathbb{R}^{N\times K\times d}$;
  % The validness indicator $\mathbf{V}\in \mathbb{R}^{Q\times K}$ for the validness of the stored features in  $\mathbf{M}^{s}$ and $\mathbf{M}^{t}$;
  % The maximum valid steps $U$;
% \State Randomly initialize $\mathcal{E}$, $\mathcal{C}$, and $\mathbf{S}$;
% \State Zero initialize $\mathbf{V}$;
\For{the $i$-th iteration in training}
% \State Get each sample in current iteration;
% \For{each sample in the current iteration}
% \State Get features $\{\mathbf{f}^t_i\}_{i=1}^{N}$ and $\{\mathbf{f}^s_i\}_{i=1}^{N}$  by  $\mathcal{T}$ and $\mathcal{S}$;
% $\{\mathbf{f}^t_i\}_{i=1}^{L}$;
% \State Get features $\{\mathbf{f}^s_i\}_{i=1}^{N}$ extracted by $\mathcal{S}$;
% \State Calculate $\mathcal{L}_{fcd}$ of $\{\mathbf{f}^s_i\}_{i=1}^{N}$ and $\{\mathbf{f}^t_i\}_{i=1}^{N}$ by Eq.~(\ref{fcd});
% \State  features $\{f_i\}_{i=1}^{m} = \mathcal{E}(B)$;
% \For{features $\mathbf{f}^t_i$ and $\mathbf{f}^s_i$ in $\{\mathbf{f}^t_i\}_{i=1}^{N}$ and $\{\mathbf{f}^s_i\}_{i=1}^{N}$}
\State Set the scale $g_i$ based on Eq.~\ref{eq:scale};
\State Set the  position offset $t_i$ based on Eq.~\ref{eq:augt};
% \State 
\State Modify the RoPE position embeddings based on Eq.~\ref{pi3};
\State Train model $\mathcal{H}$ on training window $R$;
\State Compute the next token prediction loss;
% for $\mathcal{H}$.
% based on $\mathbf{V}$ and the corresponding label $y_i$;
% =\mathop{\mathrm{argmin}}\limits_{j=1:K}
% \mathbf{V}[y_i][j]$
% \State $idx_i = \mathop{argmin}\limits_{j=1:K}(\mathbf{V}[y_i][j])$
\State Update parameters of model $\mathcal{H}$;
% \return
\EndFor
% \EndFor
\Ensure
  The optimized long context model $\mathcal{H}'$. (Note that $\mathcal{H}'$ can extend to different context windows at inference.);
\end{algorithmic}
% \vspace{-2mm}
\end{algorithm}
\subsubsection{Training and Inference}
\label{train}
As discussed in Sec.~\ref{sec:intro},
the training procedure is performed once for our E$^{2}$-LLM,
and we can extend to different evaluation context windows easily at inference. The details are as follows.

\noindent\textbf{Training}.
In training,
% for each sample,
first,
for the $i$-th iteration,
based on  $g_i$ and position offset $t_i$ in training,
we replace the $g$ and $t$  with $g_i$ and $t_i$ for Eg.~\ref{pi3},
respectively.
% following Eq.~\ref{pi3},
% we  modified RoPE is written as follows:
% our {E$^{2}$-LLM will allow our $\mathcal{H}$ perform well on extreme length extension.} 
Then,
we fine-tune the LLM $\mathcal{H}$ with a short context window $R$ using the next token prediction task with modified position encodings on the trained context window.
For better clarification, we also provide an algorithm of our proposed E$^{2}$-LLM method in Alg.~\ref{alg:pseudocode}.
% we directly use the supervised fine-tuning strategy.

\noindent\textbf{Inference}.
% Following PI~\cite{position-interpolation}
Our E$^{2}$-LLM also does not introduce extra training weights, or modify the network architecture in any way,
which means that it is attractive in practical applications as most infrastructure and optimization for the original model can be reused after length extension.
At inference,
we can extend to different context windows by setting different scale parameters for interpolation easily.
For example,
we set $g=8$ for interpolating to $32,768$ and $g=16$ for interpolating to $65,536$,
which are called as E$^{2}$-LLM-32k and E$^{2}$-LLM-64k, respectively.
It should be mentioned that the weights of E$^{2}$-LLM-32k and E$^{2}$-LLM-64k are the same at inference, and the only difference is that the scale parameters are set as 8 and 16, respectively.
Moreover,
in practice,
we can only deploy one LLM on devices,
and automatically change the scale parameter of RoPE based on the length of input context to support different context windows.

% \section{Experiments}

\begin{table*}[t]
\setlength\tabcolsep{4pt} 
\caption{Results (\%) on single-doc QA, multi-doc QA and summarization tasks from LongBench dataset.}
\centering  
\resizebox{\textwidth}{!}{
\begin{tabular}{l|cccc|cccc|cc}
\toprule
\multirow{2}{*}{\textbf{Model}} & \multicolumn{4}{c|}{\textbf{Single-Doc QA}} & \multicolumn{4}{c|}{\textbf{Multi-Doc QA}} & \multicolumn{2}{c}{\textbf{Summarization}} \\
\cmidrule(lr){2-5} \cmidrule(lr){6-9} \cmidrule(lr){10-11} 
% & \textbf{1-1} & \textbf{1-2} & \textbf{1-3} & \textbf{1-4} & \textbf{2-1} & \textbf{2-2} & \textbf{2-3} & \textbf{2-4} & \textbf{3-1} & \textbf{3-2} & \textbf{3-3} & \textbf{3-4} \\
 % \begin{tabular}[c]{@{}c@{}}Mobile\\ Latency\end{tabular} 
&\begin{tabular}[c]{@{}c@{}}\textbf{Narrative}\\ \textbf{QA}\end{tabular}& \textbf{Qasper} &\begin{tabular}[c]{@{}c@{}}\textbf{MultiField}\\ \textbf{QA-en}\end{tabular}  & \begin{tabular}[c]{@{}c@{}}\textbf{MultiField}\\ \textbf{QA-zh}\end{tabular}  & \begin{tabular}[c]{@{}c@{}}\textbf{Hotpot}\\ \textbf{QA}\end{tabular} & 
\begin{tabular}[c]{@{}c@{}}\textbf{2WikiMulti}\\ \textbf{hopQA}\end{tabular} &\begin{tabular}[c]{@{}c@{}}\textbf{MuSi}\\ \textbf{Que}\end{tabular} & 
\begin{tabular}[c]{@{}c@{}}\textbf{Du}\\ \textbf{Reader}\end{tabular}& \begin{tabular}[c]{@{}c@{}}\textbf{Gov}\\ \textbf{Report}\end{tabular}& \textbf{QMSum} \\
\midrule
GPT-3.5-Turbo-16k & 23.6 & 43.3 & 52.3 & 61.2 & 51.6 & 37.7 & 26.9 & 28.7 & 29.5 & 23.4  \\
\midrule

% \rowcolor{mygray}
Llama2-7B-chat-4k & 18.7 & 19.2 & 36.8 & 11.9 & 25.4 & 32.8 & 9.4 & 5.2 & 27.3 & 20.8  \\
LongChat-v1.5-7B-32k & 16.9 & 27.7 & 41.4 & 29.1 & 31.5 & 20.6 & 9.7 & 19.5 & 30.8 & 22.7\\
% \rowcolor{mygray}XGen-7B-8k & 18.0 & 18.31.51 & 37.7 & 14.8 & 29.7 & 21.1 & 10.3 & 11.0 & 27.3 & 20.5 & 26.2 & 2.2 \\
% InternLM-7B-8k & 12.1 & 16.7 & 23.4 & 33.6 & 28.7 & 22.8 & 9.0 & 11.1 & 9.7 & 15.9 & 22.8 & 12.4 \\
% \rowcolor{mygray}
% ChatGLM2-6B & 11.8 & 22.5 & 35.0 & 33.2 & 22.4 & 20.1 & 6.1 & 16.3 & 23.2 & 21.1 & 25.2 & 14.5 \\
% ChatGLM2-6B-32k & 21.1 & 31.5 & 46.2 & 51.6 & 45.1 & 34.0 & 21.9 & 37.6 & 32.4 & 24.0 & 26.5 & 16.2 \\
% ChatGLM3-6B-32k & 21.1 & 31.5 & 46.2 & 51.6 & 45.1 & 34.0 & 21.9 & 37.6 & 32.4 & 24.0 & 26.5 & 16.2 \\
% Longchat-7b-1.5-32k& 21.1 & 31.5 & 46.2 & 51.6 & 45.1 & 34.0 & 21.9 & 37.6 & 32.4 & 24.0 & 26.5 & 16.2 \\
% Llama2-7B-NTK-16k & 15.13 & 33.61 & 30.61 & 51.6 & 45.1 & 34.0 & 21.9 & 37.6 & 32.4 & 24.0 & 26.5 & 16.2 \\
Vicuna-v1.5-7B-16k & 19.4 & 26.1 & 38.5 & 43.0 & 25.3 & 20.8 & 9.8 & 19.3 & 27.9 & 22.8  \\
LongLora-7B-16k& 19.8& 29.1 & 37.2 & 8.5 & 37.0 & 30.3 & 17.1 & 15.3 & 31.5 & 24.1 \\
% LongLoea 7B
% \textbf{E$^{2}$-LLM-Llama2-7B-16k} & 16.44 & 34.74 & 39.1 & 43.61 & 37.09 & 34.43 & 17.86 & 18.62 & 29.39 & 23.01 \\

% \textbf{E$^{2}$-LLM-Llama2-7B-32k} & 12.28 & 35.56 & 40.41 & 46.63 & 43.71 & 34.82 & 21.97 & 22.56 & 29.67 & 23.75 \\
\textbf{E$^{2}$-LLM-Llama2-7B-16k} & 16.4 & 34.7 & 39.1 & 43.6 & 37.1 & 34.4 & 17.9 & 18.6 & 29.4 & 23.0 \\

\textbf{E$^{2}$-LLM-Llama2-7B-32k} & 12.3 & 35.6 & 40.4 & 46.6 & 43.7 & 34.8 & 22.0 & 22.6 & 29.7 & 23.8 \\
\midrule

Llama2-13B-chat-4k& 19.2 &25.8  & 36.9 &33.3  & 36.1 &32.4 & 14.5 & 26.8 &26.6 & 20.2 \\

% Yarn-13B-128k & 13.28 & 8.06 & 19.79 & 19.26 & 10.37 & 10.06 & 5.53 & 17.14 & 23.38 & 19.25 & 19.87 & 11.57 \\
Vicuna-v1.5-13B-16k & 18.9 & 29.9 & 46.2 & 28.4 & 38.1 & 36.0 & 10.7 & 20.9 & 27.9 & 22.1  \\
% PI-Llama2-13B-16k& 18.32 & 31.55 & 43.58 & 27.04 & 36.7 & 34.7 & 11.89 & 21.32 & 27.36 & 22.15 & -& 15.74 \\
% \textbf{E$^{2}$-LLM-LLama2-7B-32k} & 24.86 & 40.48 & 42.98 & 42.69 & 46.3& 41.53 & 26.03 & 21.27 & 29.91 & 23.48 & - & 9.67 \\
% \textbf{E$^{2}$-LLM-LLama2-13B-16k} & 25.44 & 34.83 & 42.77 & 40.72 & 46.28& 36.33 & 19.31 & 11.76 & 29.4 & 23.26 & 25.2 & 9.79 \\
% \textbf{E$^{2}$-LLM-LLama2-13B-32k} & 22.77 & 36.86 & 44.61 & 43.92 & 46.21& 37.28 & 21.3 & 12.06 & 29.43 & 24.26 & 25.28 & 10.75 \\
{PI-Llama2-13B-16k} & 19.2 & 33.3 & 42.7 & 47.9 & 44.9& 34.8 & 19.5 & 17.4 & 27.9 & 23.7 \\
\textbf{E$^{2}$-LLM-Llama2-13B-16k} & 25.4 & 35.3 & 46.5 & 49.1 & 46.4& 38.3 & 25.2 & 19.3 & 29.9 & 22.7  \\
\textbf{E$^{2}$-LLM-Llama2-13B-32k} & 24.1 & 36.2 & 49.0 & 52.5 & 49.2& 37.6 & 23.1 & 20.4 & 29.9 & 23.1 \\
% \textbf{E$^{2}$-LLM-LLama2-13B-32k} & 25.44 & 34.83 & 42.77 & 40.72 & 46.28& 36.33 & 19.31 & 11.76 & 29.4 & 23.26 & 25.2 & 9.79 \\

% \textbf{E$^{2}$-LLM-LLama2-13-32k} & 25.44 & 26.1 & 38.5 & 43.0 & 25.3 & 20.8 & 9.8 & 19.3 & 27.9 & 22.8 & 27.2 & 15.1 \\
\bottomrule
\end{tabular}
}
\label{tb:exp1}
\end{table*}

% \rowcolor{mygray}

\begin{table*}[t]
\setlength\tabcolsep{4pt} 
\caption{Results (\%) on summarization, few-shot learning, synthetic, and code tasks from  LongBench dataset. `Overall' is computed by the macro-average over major task categories. This is computed on English (EN) tasks, Chinese (ZH) tasks, and all (All) tasks, code tasks are included in both languages.}
\centering
\resizebox{1.0\textwidth}{!}{
\begin{tabular}{l|cc|cccc|cc|ccc}
\toprule
\multirow{2}{*}{\textbf{Model}}& \multicolumn{2}{c|}{\textbf{Summarization}} & \multicolumn{4}{c|}{\textbf{Few-shot Learning}}  & \multicolumn{2}{c|}{\textbf{Code}} & \multicolumn{3}{c}{\textbf{Overall}} \\
\cmidrule(lr){2-3}
\cmidrule(lr){4-7} \cmidrule(lr){8-9} \cmidrule(lr){10-12}
% & \textbf{4-1} & \textbf{4-2} & \textbf{4-3} & \textbf{4-4} & \textbf{5-1} & \textbf{5-2} & \textbf{5-3} & \textbf{6-1} & \textbf{6-2} & \textbf{EN} & \textbf{ZH} & \textbf{All} \\
&\textbf{MultiNews}&\textbf{VCSUM}& \textbf{TREC} & \textbf{TriviaQA} & \textbf{SAMSum} & \textbf{LSHT} & \textbf{LCC} & \textbf{RepoBench-P} & \textbf{EN} & \textbf{ZH} & \textbf{All} \\
\midrule
GPT-3.5-Turbo-16k &26.7&16.0& 68.0 & 91.4 & 41.7 & 29.2 & 54.7 & 53.6 &44.60 & 33.78& 42.19\\
\midrule
Llama2-7B-chat-4k &25.8&0.2& 61.5 & 77.8 & 40.7 & 19.8 &  52.4 & 43.8 & 35.17& 15.45 & 20.79 \\
LongChat-v1.5-7B-32k &26.4&9.9& 63.5 & 82.3 & 34.2 & 23.2  & 53.0 & 55.3 & 36.86 & 20.43 & 33.21 \\
% XGen-7B-8k & 65.5 & 77.8 & 25.3 & 20.5 & 2.1 & 8.5 & 3.5 & 38.6 & 38.6 & 28.3 & 15.1 & 25.0 \\
% InternLM-7B-8k & 52.0 & 77.8 & 21.2 & 15.2 & 3.0 & 6.0 & 0.9 & 44.1 & 28.8 & 24.2 & 18.3 & 22.6 \\
% ChatGLM2-6B & 44.5 & 70.6 & 29.5 & 20.8 & 2.5 & 3.0 & 6.5 & 49.0 & 43.2 & 26.6 & 22.9 & 25.7 \\
% ChatGLM2-6B-32k & 62.5 & 78.7 & 36.3 & 27.7 & 1.5 & 77.0 & 64.5 & 55.6 & 49.9 & 40.9 & 41.7 & 41.4 \\
Vicuna-v1.5-7B-16k &27.2&15.1& 74.0 & 86.2 & 40.8& 28.8  & 51.0& 43.5 & 36.49 & 26.55& 34.28 \\
LongLora-7B-16k & 27.7&0.5&63.5 & 85.7 & 41.9 & 26.0 & 57.6 & 54.5 & 39.79 & 14.55 & 34.18\\

\textbf{E$^{2}$-LLM-Llama2-7B-16k} &25.9&9.6& 68.5& 89.2 & 38.2 & 35.0 & 65.8& 58.1 & 41.26& 26.70 & 38.03 \\
% 25.87, 29.34
\textbf{E$^{2}$-LLM-Llama2-7B-32k} &25.4&11.7& 70.5 & 88.4& 32.5 & 40.0 & 64.5 & 60.9 & \textbf{41.74}& \textbf{30.23} & \textbf{39.18} \\
\midrule
Llama2-13B-chat-4k &26.1&17.2& 66.0&85.2&36.5&20.3&51.9&52.8 &37.87&24.38&34.87\\
% Yarn-13B-128k & 26.13 & 20.48 & 19.62 & 28.4 & 50.15 & 47.25 & - & - & - \\
Vicuna-v1.5-13B-16k &27.1&16.4& 74.0 & 84.9& 27.8 & 29.8 & 44.1& 45.6 & 38.08 & 23.86& 34.92 \\
% \textbf{E$^{2}$-LLM-LLama2-7B-32k} & 24.86 & 40.48 & 42.98 & 42.69 & 46.3& 41.53 & 26.03 & 21.27 & 29.91 & 23.48 & - & 9.67 \\
% \textbf{E$^{2}$-LLM-LLama2-13B-16k} & 74.0& 88.03 & 39.81&35.0 & -& -& - & 62.57 & 52.42& 39.16 & 20.76 & 36.10\\
% \textbf{E$^{2}$-LLM-LLama2-13B-32k} & 76.0& 89.38 & 40.03 & 44.5 & -& -& - & 60.42 & 54.16& 39.16 & 20.76 & 36.10\\
{PI-Llama2-13B-16k} &25.9&9.1& 72.5& 86.5& 27.9&31.0&  62.5 & 51.1& 40.88 & 26.35& 37.65\\
\textbf{E$^{2}$-LLM-Llama2-13B-16k} &27.0&9.8& 73.5& 87.9 & 40.6&36.0 &  65.4 & 59.1& \textbf{44.73} & 28.56& 41.13\\
\textbf{E$^{2}$-LLM-Llama2-13B-32k} &26.8&10.2& 75.0& 87.8 & 40.9 & 44.5 &  63.8 & 57.5& 44.55 & \textbf{31.93} & \textbf{41.74}\\
\bottomrule
\end{tabular}
}
\label{tb:exp2}
\end{table*}

\section{Experiments}
% \jh{Experiments are fake now, please ignore.}
\subsection{Experimental Settings}
\label{exp:setting}
\paragraph{Models.}
In our E$^{2}$-LLM,
we take the pre-trained 7B, 13B Llama2~\citep{llama2} models to demonstrate the effectiveness of our E$^{2}$-LLM.
% The maximum extended context window sizes are up to 100k for 7B models, 64k for 13B models, and 32k for 70B models. The position indices for these models are re-scaled with Position Interpolation~\citep{position-interpolation}. 

\paragraph{Training Procedure.}
All models are fine-tuned via the next token prediction objective based on two 8$\times$ A100 GPU machines. We use AdamW~\citep{adamw} with $\beta_1 = 0.9$ and $\beta_2 = 0.95$. The learning rate is set to $1\times 10^{-5}$ for 7B and 13B models, and the whole training step is set to 30,000 with a global batch size of 16.

\paragraph{Datasets.}
% We use the Redpajama~\citep{together2023redpajama} dataset for training.
The training dataset includes the pretrain dataset (i.e., Pile~\cite{gao2020pile}), 
and fine-tuning datasets (i.e., ShareGPT~\cite{zheng2023judging} and the long summarization datasets~\cite{cohan-etal-2018-discourse}).
Note that the fine-tuning datasets are used to improve the question-answer abilities of long-context LLMs following Vicuna and LongChat models~\cite{zheng2023judging} and generate reasonable results on LongBench.
% For samples from fine-tuning dataset,
% we
We evaluate the long-sequence language modeling performance of our fine-tuned models on the LongBench~\cite{bai2023longbench} and the arxiv proof-pile dataset~\cite{proof-pile}.
% /We use the test split of PG19~\citep{pg19}, consisting of 100 documents.
% For LLaMA2,
% we compare the datasets in English.
% For the proof-pile dataset, we also use the test split of it for evaluation. We follow Position Interpolation~\citep{position-interpolation} for Proof-pile data processing. We evaluate perplexity by using a sliding window approach with $S=256$, following~\citep{alibi}. 

% In addition, we build a long context QA dataset, LongQA, for supervised fine-tuning. Although the models fine-tuned with Redpajama~\citep{together2023redpajama} present good perplexities, their chat ability is limited. We collect more than 3k question-answer pairs, relating to the materials like technical paper, science fiction, and other books. The questions we designed include summarization, relationships, characters, and other details related to the material. For more details, please refer to the appendix.

\subsection{Results on LongBench}
We evaluate several popular LLMs with long context capability, including GPT-3.5-Turbo-16k~\citep{chatgpt}, Llama2-7B-chat-4k~\citep{llama2}, LongChat-v1.5-7B-32k~\citep{longchat2023}, Vicuna-v1.5-7B-16k~\citep{zheng2023judging}, Longlora-7B-16k~\cite{longlora}, Llama2-13B-chat-4k~\citep{llama2}, Vicuna-v1.5-13B-16k~\citep{zheng2023judging}, PI-Llama2-13B-16k.
LongChat-v1.5-7B-32k, Vicuna-v1.5-7B-16k, and LongLora-7B-16k are fine-tuned from Llama2-7B based on PI.
Vicuna-v1.5-13B-16k~\citep{zheng2023judging}, PI-Llama2-13B-16k are fine-tuned with Llama2-13B based on PI,
where PI-Llama2-13B-16k are fine-tuned with our constructed datasets.
Following LongBench~\cite{bai2023longbench},
we conduct the assessment in a zero-shot setting, except for the few-shot learning tasks where the few-shot examples are provided as part of the long context.
% The input format prompt and the maximum output length we used during evaluation can be found in the Appendix.
When the input length $I$ surpasses the maximum context length $L'$ of a model (indicated by the suffix of its name), we truncate the input sequence $S$ from the middle since the front and end of the sequence may contain crucial information such as the instruction or question:
$S_{1:I}\rightarrow [S_{1:\lfloor L'/2\rfloor}; S_{I-\lfloor L'/2\rfloor-1:I}]$.
% During generation, we use greedy decoding for reproducibility.
The metric for each dataset is shown in Table~\ref{tb:stat} from the Appendix~\ref{appendix:longbench}. 
% For tasks built based on previous datasets, the metrics we used are consistent with those used in the original work.
% F1 and ROUGE-L~\citep{lin2004rouge} are two popular N-gram based metrics widely adopted in QA and summarization tasks.
% Edit Sim (Levenshtein distance) is popularly used in code generation evaluation~\citep{svyatkovskiy2020intellicode}.
% For the few-shot learning tasks, we extract the first line of the response.
% For the two code completion tasks, we extract the first line of model generation that is not comment.

% \begin{figure}[t]
%     \centering
%     \includegraphics[width=\textwidth]{figs/radar_new.pdf}
%     \caption{Average scores on 6 major tasks, on English and Chinese datasets, respectively.}
%     \label{fig:radar}
% \end{figure}

% \noindent\textbf{Results on LongBench}
As shown in Table~\ref{tb:exp1} and Table~\ref{tb:exp2},
we report the performance results (\%) on the LongBench dataset. 
% Additionally, Figure~\ref{fig:radar} presents a radar plot depicting the models' abilities on the 6 major tasks.
Specifically,
the key findings from the experiment results are as follows:
(1) When compared with the commercial model (GPT-3.5-Turbo-16k) with an overall accuracy of 44.60\% in English,
our E$^{2}$-LLM-Llama2-13B-32k achieves closing results with an overall accuracy of 44.55\%.
(2) In Table~\ref{tb:exp1} and Table~\ref{tb:exp2},
we also evaluate the results of our E$^{2}$-LLM with different evaluation context sizes (i.e., 16k and 32k),
and we observe that the performance results are better when we extend the evaluation context window size.
Besides,
as the lengths of most documents in LongBench are less than 16k,
the improvements on these tasks are not significant when we increase the evaluation context window.
% \jh{For the LSHT task with an average length of 22k,
% we can see the PI-Llama2-13B-32k is better than the PI-Llama2-13B-16k a lot.}
(3) For a fair comparison,
we also reimplement the positional interpolation method based on Llama2-13B (i.e., PI-Llama2-13B-16k) with the same training strategy and training datasets.
When compared with PI-Llama2-13B-16k,
our E$^{2}$-LLM-Llama2-13B-16k still achieves significant improvements on LongBench,
which further demonstrates the effectiveness of our E$^{2}$-LLM.

\subsection{Results on Proof-Pile}
\label{sec:perplexity}
% \jh{proof-pile needed?}
On the cleaned Arxiv Math proof-pile dataset~\cite{proof-pile},
we evaluate the long sequence language modeling performance of our extended models and baseline methods (i.e., Vicuna-v1.5-16k and LongChat-v1.5-32k),
where the perplexity results on reported.
% book corpus (PG-19)\citep{pg19} and
% cleaned Arxiv Math proof-pile dataset \citep{proof-pile}. 
% We use the test splits of PG19~\citep{pg19}. 
% For PG19, we use the whole test split consisting of 100 documents. 
For the proof-pile dataset, we randomly sample 128 documents with at least 64k tokens and evaluate the calculated perplexity of each of these samples.
All perplexity evaluations were calculated using the sliding window method from~\cite{alibi} with $S=256$.
Specifically,
Vicuna-v1.5-16k and LongChat-v1.5-32k are fine-tuned on the Llama2 model and linear RoPE scaling method,
which is based on the Position Interpolation (i.e., PI)~\cite{position-interpolation}.
% which
In Table~\ref{table:perplexity-pile},
 we found that models extended with our method enjoy a significantly improved perplexity from longer context window sizes when compared with other baseline methods.
Besides,
for other methods,
the training context window is equal to the maximum evaluation context window,
thus the training costs are very large when the window size is large,
where the training costs are shown in Fig.~\ref{fig:intro}.
In contrast,
our E$^{2}$-LLM only needs to train Llama models once and the training context window is short, which reduces the training costs greatly.

\label{sec:experiments}

\begin{table*}[thbp]
\caption{Evaluation perplexity on Arxiv Proof-pile dataset~\cite{proof-pile} based on Llama2 7B and 13B models,
where lower perplexity means better performance.
``PI'' denotes Position Interpolation~\cite{position-interpolation}. The open-sourced Vicuna-v1.5-16k and LongChat-v.15-32k are extended based on the PI method. Note that the weights of E$^{2}$-LLM-16k, E$^{2}$-LLM-32k and E$^{2}$-LLM-64k are the same at inference, and the only difference is that the scale parameters are set as 4, 8 and 16, respectively.}
\centering
\begin{tabular}{cccccccc}
\toprule
\multicolumn{3}{c}{Model} & \multicolumn{5}{c}{Evaluation Context Window Size}\\
\hline
Size &  Training Context Window & Method &  4096 & 8192 & 16384 & 32768 &65536\\
% \midrule
%    7B &            4096 &   None &  7.20 &    $>10^3$ &  $>10^3$ &    $>10^3$ &     $>10^3$ \\
%    7B &            8192 &     FT &  7.21 & 7.34 & 7.69 &     - &     - \\
% \midrule
%    7B &            8192 &     PI&  7.13 & 6.96 & 6.95 &     - &     - \\
%    7B &           16384 &     PI &  7.11 & 6.93 & 6.82 &  6.83 &     - \\
%    % 7B &           32768 &     PI&  7.23 & 7.04 & 6.91 &  6.80 &  6.77 \\
%       % 7B &           8192 &     LongLora&  7.23 & 7.04 & 6.91 &  6.80 &  6.77 \\
%       % 7B &           16384 &     LongLora&  7.23 & 7.04 & 6.91 &  6.80 &  6.77 \\
%             % 7B &           8192 &     LongLora&  7.23 & 7.04 & 6.91 &  6.80 &  6.77 \\
%       % 7B &           32768 &     LongLora&  7.23 & 7.04 & 6.91 &  6.80 &  6.77 \\
%                  7B &           w/o train &     NTK&  6.54 & 6.40 & 6.28 &  6.18 &  6.09 \\ 
%                                7B &           8192 &     NTK&  6.54 & 6.40 & 6.28 &  6.18 &  6.09 \\ 
%             7B &           8192 &     Yarn&  6.54 & 6.40 & 6.28 &  6.18 &  6.09 \\
                   % 7B &           8192 &     ABF&  6.54 & 6.40 & 6.28 &  6.18 &  6.09 \\

\midrule
   7B &            4k &   None &  2.92 &    - &  - &    - &     -\\
   % 7B &            8192 &     FT &  6.56 & 6.57 & 6.69 &     - &     - \\
\midrule
   \multirow{2}{*}{7B}  &            16k &     Vicuna-v1.5-16k (PI)&  3.48 & 3.17& 3.95 &      - &     -\\
    &           32k &     LongChat-v1.5-32k (PI)&  3.54& 3.18 & 2.91&  2.73 &    - \\

\midrule
          &           \multirow{3}{*}{4k} &     E$^{2}$-LLM-16k&  2.96 & 2.71& 2.54 &     - &     - \\
   7B &            &     E$^{2}$-LLM-32k &  2.99 & 2.74 & 2.56&2.46  &     - \\
       &            &     E$^{2}$-LLM-64k &  3.06 & 2.81 & 2.62 & 2.51 &     2.56 \\
\midrule
   13B &            4k &   None &  2.78 &    - &  - &    - &     -\\
   % 13B &            8192 &     FT &  6.56 & 6.57 & 6.69 &     - &     - \\
\midrule
   13B &            16k &     Vicuna-v1.5-16k (PI)&  3.27& 2.97 & 2.78 &     - &     - \\
   % 13B &           32k &     PI &  - & - & -&  - &     - \\
   % 13B &           16k &     NTK&  6.54 & 6.40 & 6.28 &  6.18 &  6.09 \\ 
   % 13B &           32768 &     PI&  6.54 & 6.40 & 6.28 &  6.18 &  6.09 \\
      % 13B &            32k &     LongLora-32k&  6.55 & 6.42 & 6.42 &     - &     - \\
   % 13B &           16384 &      LongLora &  6.56 & 6.42 & 6.31 &  6.32 &     - \\
   % 13B &           64k &     LongLora-64k&  6.54 & 6.40 & 6.28 &  6.18 &  6.09 \\
                    % 13B &           w/o train &     NTK&  6.54 & 6.40 & 6.28 &  6.18 &  6.09 \\ 
      % 13B &           64k &     Yarn-64k&  6.54 & 6.40 & 6.28 &  6.18 &  6.09 \\
\midrule
          &             &     E$^{2}$-LLM-16k&  2.82 & 2.59& 2.43 &     - &     - \\
   13B &           4k &     E$^{2}$-LLM-32k &  2.85 & 2.61 & 2.44 &2.34  &     - \\
       &            &     E$^{2}$-LLM-64k &  2.91 & 2.67 & 2.49 & 2.39 &     2.44 \\

             % 13B &           8192 &     ABF&  6.54 & 6.40 & 6.28 &  6.18 &  6.09 \\
                   % 13B &           8192 &     ABF&  6.54 & 6.40 & 6.28 &  6.18 &  6.09 \\
% \midrule
%   33B &            2048 &   None &  5.82 &    - &    - &     - &     - \\
%   33B &            8192 &     FT &  5.88 & 5.99 & 6.21 &     - &     - \\
% \midrule
%   33B &            8192 &     PI &  5.82 & 5.69 & 5.71 &     - &     - \\
%   33B &           16384 &     PI &  5.87 & 5.74 & 5.67 &  5.68 &     - \\
% \midrule
% 65B & 2048 & None & 5.49 & - & - & - & -\\
% \midrule
% 65B & 8192 & PI & 5.42 & 5.32 & 5.37 & - & - \\  
\bottomrule
\end{tabular}
\label{table:perplexity-pile}
\end{table*}

\subsection{Ablation Study}
\noindent\textbf{Effect of the augmentation strategies.}
In Table~\ref{tab:a1}
we provide two alternative variants of our E$^{2}$-LLM (i.e., E$^{2}$-LLM (w/o aug on $t$), E$^{2}$-LLM (w/o aug on $g$)) to train the LLama2-13B base model.
Specifically, 
for E$^{2}$-LLM (w/o aug on $t$), we only use the augmentation on the scale parameter $g$ without using the augmentation on the position offset $t$,
and we evaluate the performance by extending the context window as 32k. For E$^{2}$-LLM (w/o aug on $g$), we only use the augmentation on the position offset $t$ and fix the scale parameter $g$ as $2$ with the training context window of 8k in our E$^{2}$-LLM. 
Note that the evaluation context window of E$^{2}$-LLM (w/o aug on $g$) is also set as 8k.
As shown in Table~\ref{tab:a1}, our E$^{2}$-LLM is better than these two alternative variants on LongBench, which shows that it is beneficial to perform the augmentation on $t$ and $g$.
\begin{table}[!htp]
    \centering
    \caption{Ablation on different augmentation strategies.}
    % \vspace{-3mm}
            % \resizebox{0.8\linewidth}{!}
            {
    \begin{tabular}{c|ccc}  
    \toprule
   {Methods}  & EN & ZH&All\\
    \midrule
         E$^{2}$-LLM&\textbf{44.55}&\textbf{31.93}&\textbf{41.74}\\
         % LTA-PCS (w/o inter \& w/o intra)&87.8&88.2\\
      E$^{2}$-LLM (w/o aug on $t$)&42.28&29.49&39.44\\
          E$^{2}$-LLM (w/o aug on $g$)&41.66&28.33&38.77\\
           
    \bottomrule
    \end{tabular}}
    % \vspace{-4mm}
    \label{tab:a1}
    \end{table}
% \vspace{-5mm}

\begin{figure}[t]
\begin{center}
\includegraphics[width=0.9\linewidth]{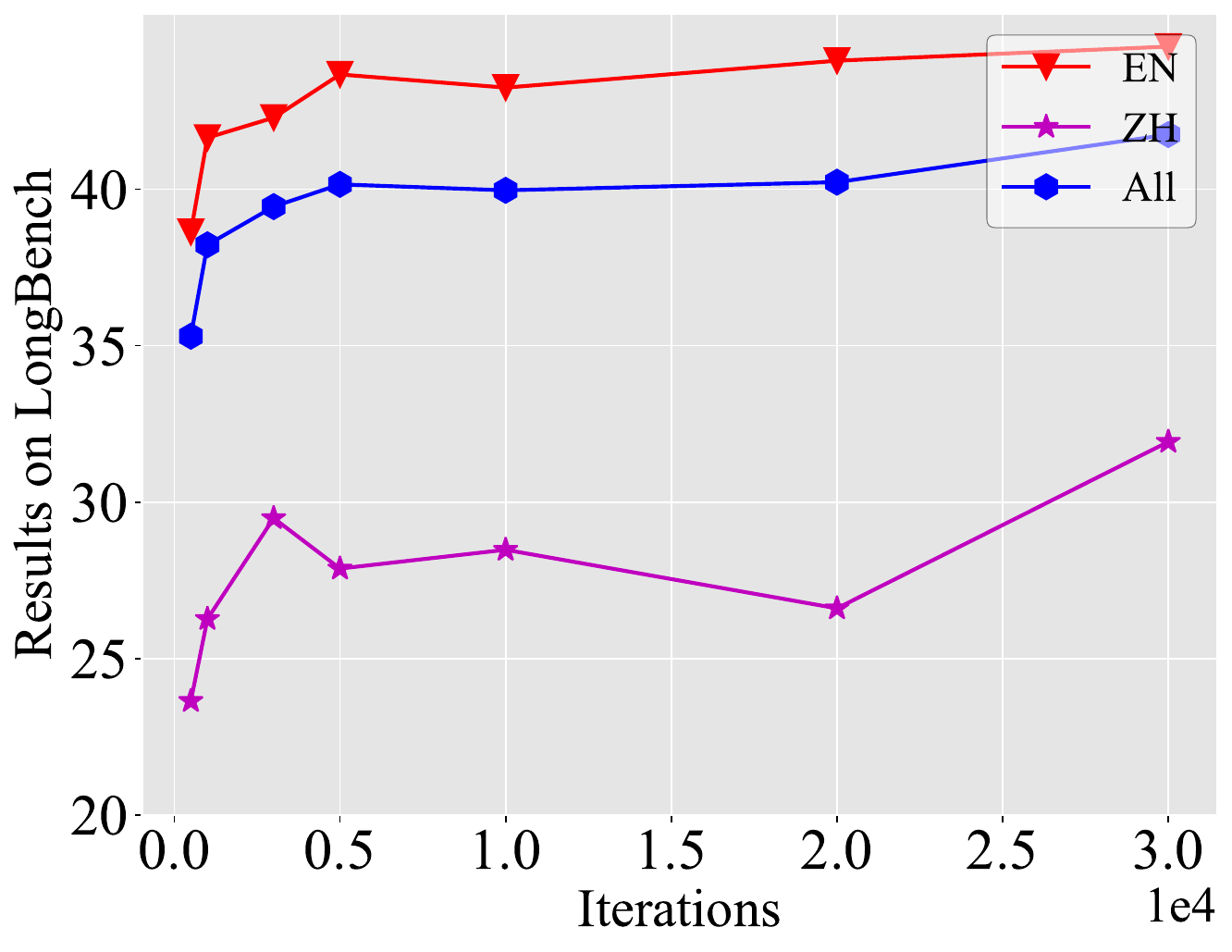}
% \vspace{-4mm}
\caption{Performance results on the Longbench dataset when increasing the training steps.
}
% \vspace{-4mm}
\label{fig:iter}
\end{center}
\end{figure}

\noindent\textbf{Effect of the number of finetuning steps.}
As shown in Fig~\ref{fig:iter},
we report the relationship between the results on the LongBench dataset and the number of fine-tuning steps for the LLaMA2 13B model using E$^{2}$-LLM,
where the results are reported on the evaluation window size of 32k.
In Fig.~\ref{fig:iter},
% in the tuning process,
% we observe that the performance results are better.
% Specifically,
at the first 5k iterations,
the results improve quickly,
which indicates that the models can gain long-context understanding abilities without a long training process.
Furthermore,
when increasing the training iterations, we observe that 
the performance results of LongBench can still increase steadily after 5k iterations.  

% are better
% the of effectively using sequences longer  than the pre-training settings.
% We can see without fine-tuning (at step 0) the model can exhibit certain language modeling capability, as 
% indicated by $<20$ perplexity for extending to 8192 context window (in contrast, the direct extrapolation method leads to $>10^3$ perplexity).
% With fine-tuning, we observed that the perplexity improves quickly. 
% At 200 steps the models surpassed the original model's perplexity on 2048 context window size, indicating the models
% gaining ability of effectively using sequences longer  than the pre-training settings for language modeling.
% At 1000 steps, we can see the models have improved steadily and achieve a significantly better perplexity.

% \jh{including tuning windows,}

\noindent\textbf{Effect of the maximum scale parameter $G_{max}$.}
In Table~\ref{tab:num_scale}, on LongBench dataset,
we also report the results of our E$^{2}$-LLM based on the Llama2-13B model to analyze the effect of the maximum scale parameter $G_{max}$ in training,
and the evaluation context window is set as 32k.
When $G_{max}$  increases from 5 to 20, the performance results on Longbench are better.
% can achieve better performance results, 
which indicates that it is effective to cover different densities by using a large $G_{max}$ in our E$^{2}$-LLM.
However, when we continue to increase the maximum scale parameter $G_{max}$, the performance improvement becomes relatively stable on LongBench.
Thus,
we directly set $G_{max}$ as 20 to support a maximum extension window of 80k.
% Thus, we set the projected number of views as 10 to achieve better performance with acceptable memory usage. 

    \begin{table}[!htp]
    \centering
    \caption{Ablation on the maximum scale parameter $G_{max}$.}
    \vspace{-3mm}
    \begin{tabular}{c|cccc}  
    \toprule
   { $G_{max}$}  & 5 & 10 &20&30\\
    \midrule
    % \multirow{5}{*}{Task}&
         EN&43.20&44.25&44.55&44.01\\
                  ZH&29.33&30.28&39.93&32.76\\
         All&40.12&41.15&41.74&41.51\\

    \bottomrule
    \end{tabular}
    % \vspace{-4mm}
    \label{tab:num_scale}
    \end{table}
\subsection{Further Analysis}

\begin{figure}[t]
\begin{center}
\includegraphics[width=1.0\linewidth]{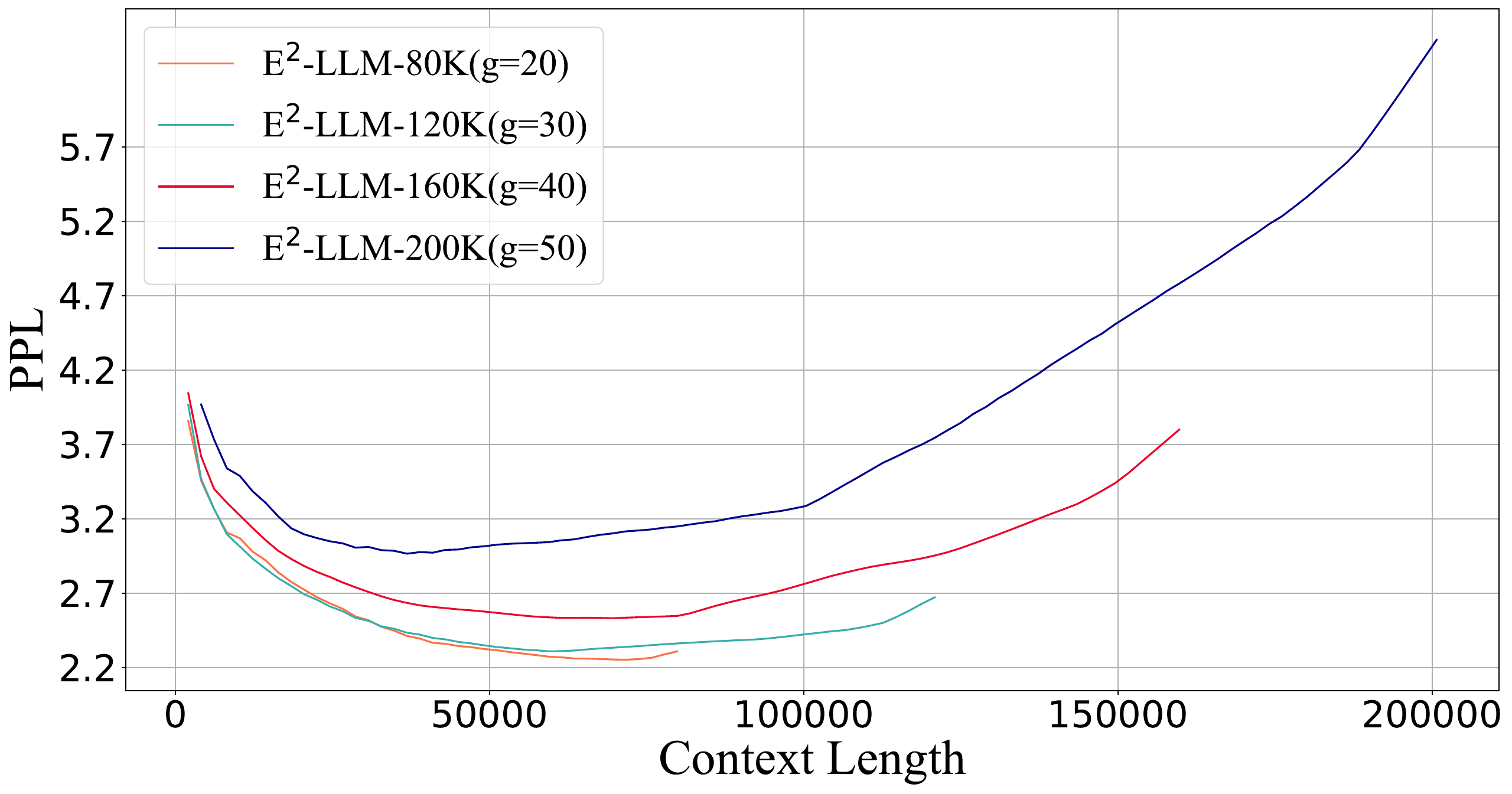}
% \vspace{-5mm}
\caption{
Generalization abilities on the unseen scales.
}
 \label{fig:vis_1a}
 % \vspace{-6mm}
\end{center}
\end{figure}

\noindent\textbf{Extension to unseen scales.}
By default, 
we set $G_{max}$ as 20 to support the maximum interpolated context window of 80K.
In Fig.~\ref{fig:vis_1a}, the interpolation scales are experimentally adjusted to 20, 30, 40, and 50 during inference to evaluate the generalization ability of E$^2$-LLM. 
The results demonstrate that PPL maintains a satisfactory level for contexts comprising fewer than 120K tokens.
Nonetheless, when we continue to increase the scale, a discernible deterioration in performance occurs. 
It suggests that E$^2$-LLM possesses robust generalization ability for unseen or OOD scales within a certain range.

\begin{figure}[t]
\begin{center}
\includegraphics[width=1.0\linewidth]{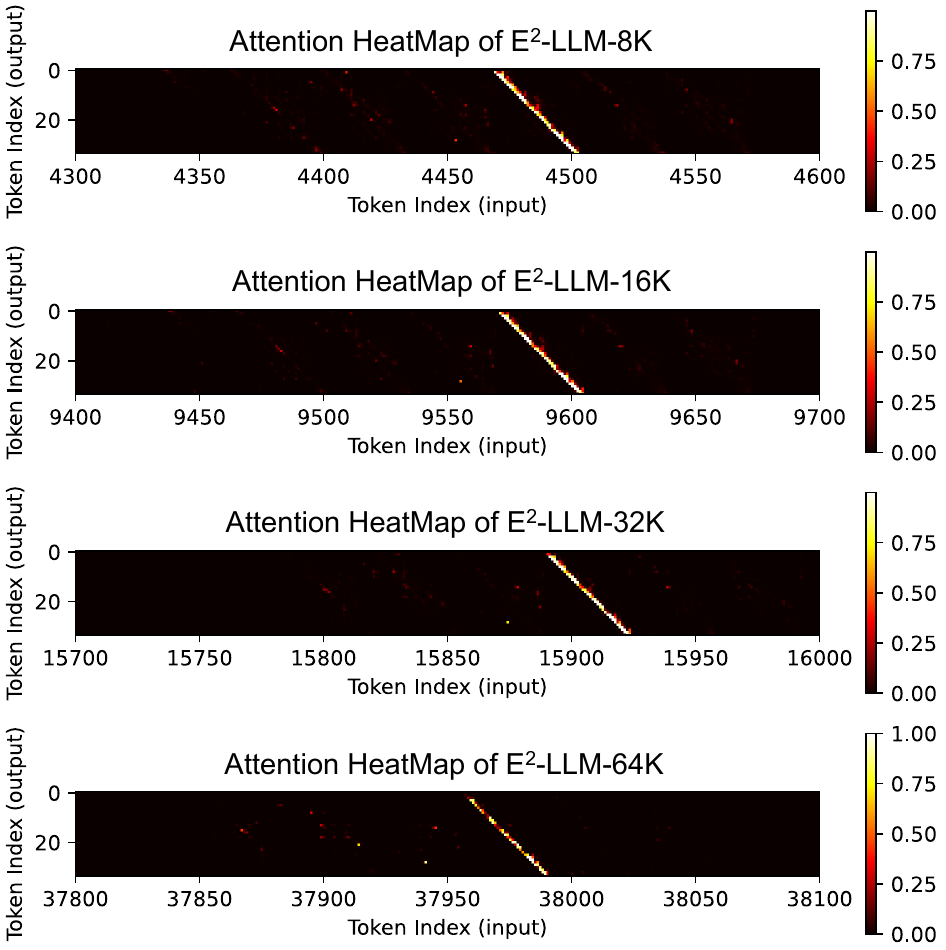}
% \vspace{-4mm}
\caption{Visualization of attention heatmaps on 8k, 16k, 32k and 64k input contexts.
}
% \vspace{-4mm}
\label{fig:heatmap}
\end{center}
\end{figure}

\noindent\textbf{Visualization on Attention Map.}
To further analyze the effect of E$^{2}$-LLM, we visualize the attention heatmaps in the layer for our E$^{2}$-LLM-8k, E$^{2}$-LLM-16k, E$^{2}$-LLM-32k and E$^{2}$-LLM-64k based on Llama2-13B in Fig.~\ref{fig:heatmap} with the evaluation context windows of 8k, 16k, 32k and 64k by setting scale parameter as 2, 4, 8, 16, respectively.
Specifically, as shown in Fig.~\ref{fig:heatmap}, the vertical coordinate represents the indices of the generated sequence's token, and the horizontal coordinate represents the indices of the input sequence's tokens. The input text is part of a long paper, which is truncated to 16k, 32k and 64k, respectively.
Then, three random key-value pairs are inserted into the input text, and a question is appended at the end of the text.
Note the random key-value pairs and question are as shown in Appendix~\ref{template-gpt-evaluate-eng},
% as to what the value corresponding to one of the keys is.
The question was answered correctly for 8k, 16k, 32k and 64k. 
In Fig.~\ref{fig:heatmap},
we visualize the attention heatmaps of the output sequence corresponding to the input. The ground-truth indices of the values corresponding to the keys asked in 8k, 16k, 32k and 64k are [4470, 4503], [9572,9605], [15891,15924] and [37958, 37991],
respectively,
and we observe that the attention values of the output sequence at these positions are very significant, which represents that E$^{2}$-LLM can well index the correct position when generating the responses.
% Meat the same time, it has a very good fitness and robustness to different lengths of text.

\section{Conclusion}
In this study, we introduce an Efficient and Extreme length extension method for LLMs, named E$^{2}$-LLM, which is designed to extend the context windows with a single training phase and minimal computational overhead. Notably, 
in our E$^{2}$-LLM, there is no requirement to amass extensive long-context datasets (\textit{e.g.}, samples with 32k or 64k tokens ) for training.
% To elaborate, our E$^{2}$-LLM technique necessitates training data of a significantly shorter length, such as 4k or 8k, which substantially reduces resource consumption, including the use of GPU memory. Furthermore, E$^{2}$-LLM employs a single training iteration on a brief context window, which is then adaptable to various evaluation context lengths.
Specifically, our E$^{2}$-LLM harnesses RoPE-based position embeddings to implement a pair of novel augmentation strategies that adjust the scale and position index parameters across different training samples with short lengths (\textit{e.g.}, 4k/8k).
 Comprehensive experimental results on multiple benchmark datasets demonstrate the effectiveness of our E$^{2}$-LLM on long-context tasks.
 \section{Future Works}
 For the future directions,
 we have three plans as follows: (1) as our E$^{2}$-LLM is efficient and effective and  we will try to apply our E$^{2}$-LLM on larger models (\textit{e.g.}, LLama2 70B) and larger context windows (\textit{e.g.}, 128k/192k); (2) we believe that our E$^{2}$-LLM is a general method and we will try to apply our E$^{2}$-LLM on more types of position encodings and more types of LLMs; (3) we will release our models and codes.
\bibliography{anthology,custom}
\bibliographystyle{acl_natbib}

\appendix
\onecolumn

\section{More Details}
\label{sec:appendix}
 \subsection{More details of the LongBench dataset}
 \label{appendix:longbench}
The details are shown in Table~\ref{tb:stat}.
% This is a section in the appendix.

\subsection{More details of the attention map visualization}
The three random key-value pairs and the input question are shown as follows.

% \onecolumn
\begin{figure*}
    % \centering
    % \includegraphics{}
    % \caption{Caption}
    % \label{fig:enter-label}
\begin{tcolorbox}
% \label{box:sample-box}
[colback=white!95!gray,colframe=gray!50!black,rounded corners,label={template-gpt-evaluate-eng}, title={Input Context}]
\textbf{Key-value pairs: }
\textcolor{black}{
Extract the value corresponding to the specified key in the JSON object below.
\{"2a8d601d-1d69-4e64-9f90-8ad825a74195": "bb3ba2a5-7de8-434b-a86e-a88bb9fa7289",
 "9f4a92b9-5f69-4725-ba1e-403f08dea695": "703a7ce5-f17f-4e6d-b895-5836ba5ec71c",
 "52a9c80c-da51-4fc9-bf70-4a4901bc2ac3": "b2f8ea3d-4b1b-49e0-a141-b9823991ebeb"\}}
 
\textbf{Question: }
\textcolor{black}{What is the value of key "9f4a92b9-5f69-4725-ba1e-403f08dea695"?}
\end{tcolorbox}
\end{figure*}

% \twocolumn
% \subsection{}
\begin{table*}[!htp]
\centering  
\caption{An overview of the LongBench dataset. 
``Source'' denotes the origin of the context. ``Avg len'' (average length) represents the mean length, which is calculated by counting words in English (code) datasets and characters in Chinese datasets. ``Accuracy (CLS)'' and ``Accuracy (EM)'' are classification accuracy and  exact match accuracy, respectively.}
\resizebox{\textwidth}{!}{
\begin{tabular}{lclrccc}
\toprule
Dataset & ID & Source & Avg len & Metric & Language & \#data \\
\midrule
\emph{Single-Document QA} \\
NarrativeQA & 1-1 & Literature, Film & 18,409 & F1 & English & 200 \\
Qasper & 1-2 & Science & 3,619 & F1 & English & 200 \\
MultiFieldQA-en & 1-3 & Multi-field & 4,559 & F1 & English & 150 \\
 MultiFieldQA-zh &  1-4 &  Multi-field &  6,701 &  F1 &  Chinese &  200 \\
\midrule
\emph{Multi-Document QA} \\
HotpotQA & 2-1 & Wikipedia & 9,151 & F1 & English & 200 \\
2WikiMultihopQA & 2-2 & Wikipedia & 4,887 & F1 & English & 200 \\
MuSiQue & 2-3 & Wikipedia & 11,214 & F1 & English & 200 \\
 DuReader &  2-4 &  Baidu Search &  15,768 &  Rouge-L &  Chinese &  200 \\
\midrule
\emph{Summarization} \\
GovReport & 3-1 & Government report & 8,734 & Rouge-L & English & 200 \\
QMSum & 3-2 & Meeting & 10,614 & Rouge-L & English & 200 \\
MultiNews & 3-3 & News & 2,113 & Rouge-L & English & 200 \\
 VCSUM &  3-4 &  Meeting &  15,380 &  Rouge-L &  Chinese &  200 \\
\midrule
\emph{Few-shot Learning} \\
TREC & 4-1 & Web question & 5,177 & Accuracy (CLS) & English & 200 \\
TriviaQA & 4-2 & Wikipedia, Web & 8,209 & F1 & English & 200 \\
SAMSum & 4-3 & Dialogue & 6,258 & Rouge-L & English & 200 \\
 LSHT &  4-4 &  News &  22,337 &  Accuracy (CLS) &  Chinese & 200 \\
\midrule
\emph{Code Completion} \\
LCC & 6-1 & Github & 1,235 & Edit Sim & Python/C\#/Java & 500 \\
RepoBench-P & 6-2 & Github repository & 4,206 & Edit Sim & Python/Java & 500 \\
\bottomrule
\end{tabular}
}
\label{tb:stat}
\end{table*}

\subsection{More results on short-length datasets}
\label{sec:llama}
We evaluate the models extended by E$^{2}$-LLM on several standard short-length benchmark tasks (i.e., PIQA~\cite{Bisk2020}, WSC~\cite{wsc}, HellaSwag~\cite{zellers2019hellaswag}, SIQA~\cite{siqa}, WinoGrande~\cite{ai2:winogrande}, RACE~\cite{lai2017large}, and NaturalQuestions~\cite{47761}) within the original context window size of 4,096,
where the zero-shot performance results are reported in Table~\ref{tab:llama_original}.
Specifically,
we report the results of extended models (E$^{2}$-LLM-32K and E$^{2}$-LLM-64K) on Llama2-7B and Llama2-13B models,
respectively.
From the results of Table~\ref{tab:llama_original},
when compared with the baseline Llama2 models,
we observe that the extended Llama2 models (\textit{e.g.}, E$^{2}$-LLM-32K, E$^{2}$-LLM-64K) still preserve comparable or even better performance results on these short-length benchmark datasets,
which further demonstrates the effectiveness of E$^{2}$-LLM on the short-context interpretation.

\begin{table*}[tp]
    \caption{Performance on a subset of general benchmarks.}
    \centering
    \resizebox{\linewidth}{!}{
    \begin{tabular}{cccccccccc}
       \toprule
       Model& PIQA& WSC& HellaSwag & SIQA & WinoGrande&Race-H & Race-M & NaturalQuestions&\textbf{Avg.} \\
       \midrule
       Llama2-7B (4K) &78.1  &67.3 &73.0 & 48.1 & 69.5 & 40.2 & 37.1 &16.4&53.7\\
       % \midrule
       % PI-16k &- & &69.8   & 77.6 & 53.3 & 40.9  & 67.8 \\
                     % PI-32k & - & &64.7   & 77.2 & 50.1 & 39.6  & 66.9 \\
                           % Yarn-64k    & - & &64.7   & 77.2 & 50.1 & 39.6  & 66.9 \\
              % E$^{2}$-LLM-16k & - & &64.7   & 77.2 & 50.1 & 39.6  & 66.9 \\
                            E$^{2}$-LLM-32K  &77.6  & 66.4&71.9    & 45.9  & 69.5 & 47.6 & 40.9 &18.8&54.8\\
                                 E$^{2}$-LLM-64K  & 77.2 &67.3 &70.9    &   46.5& 69.4 & 44.4 & 38.8 &17.5&54.0\\

       \midrule
       Llama2-13B (4K)   &78.9 &64.4    &75.7 & 51.7 & 73.5 & 63.0 & 58.9 & 20.2&60.8\\
       % \midrule
       % PI-16k &- & &69.8   & 77.6 & 53.3 & 40.9  & 67.8 \\
                     % PI-32k  & - & &64.7   & 77.2 & 50.1 & 39.6  & 66.9 \\
                           % Yarn-64k   & - & &64.7   & 77.2 & 50.1 & 39.6  & 66.9 \\
              % E$^{2}$-LLM-16k & - & &64.7   & 77.2 & 50.1 & 39.6  & 66.9 \\
                            E$^{2}$-LLM-32K&79.3 &    67.3& 75.6&  61.3&74.4 & 67.3 & 59.8 & 26.8&64.0\\
                               E$^{2}$-LLM-64K &78.8 &67.3   & 75.0 &61.1&74.5& 61.7 & 55.8 & 25.2&62.4\\
       % 33B & 8192 &  -& &80.2   & 80.7 & 60.2 & 45.7 & 75.9 \\       
       \bottomrule
    \end{tabular}
    }
    \label{tab:llama_original}
    %Arc-c, helawage, delete race-m&ceval
\end{table*}

\subsection{Extension on our E$^{2}$-LLM}
\begin{figure*}[t]
\begin{center}
\includegraphics[width=1.0\linewidth]{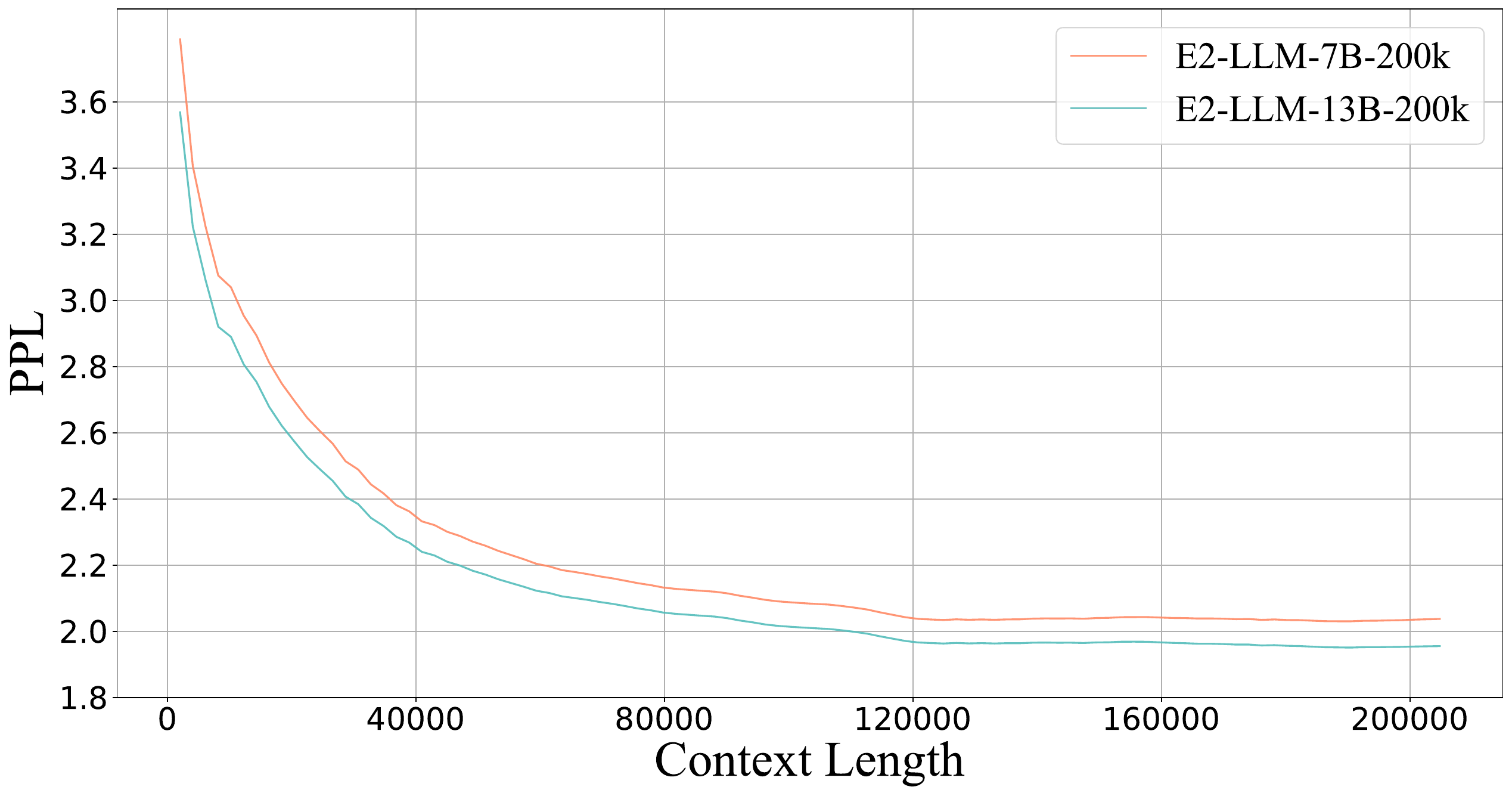}
% \vspace{-4mm}
\caption{Results of E$^{2}$-LLM on 200K context window.
}
% \vspace{-4mm}
\label{fig:200k}
\end{center}
\end{figure*}

Recently, we have updated the methodology and training strategies of our E$^{2}$-LLM,
and successfully extended the Llama2-7B and 13B to 200k context windows with dramatically lower training costs as shown in Fig.~\ref{fig:200k}.
In Fig.~\ref{fig:200k},
we report the PPL results on several documents truncated to 200k from Proof-Pile,
and we observe that the PPL results decrease smoothly,
which further demonstrate the effectiveness of our E$^{2}$-LLM.
We will add more implementation details on our newly updated E$^{2}$-LLM method in the future.

\end{document}